\documentclass{article}
\usepackage{spconf,amsmath,graphicx,hyperref}

\usepackage{booktabs}
\usepackage{multirow}

\usepackage{amssymb}
\usepackage{float}
\usepackage{caption}
\usepackage{subcaption}
\usepackage{xcolor}
\usepackage{hyperref}
\usepackage{enumitem}
\usepackage{mathtools}

\usepackage{siunitx}

\usepackage[normalem]{ulem}

\def\x{{\mathbf x}}

\def \enc { {\bbPhi_{\rmE}} }
\def \dec { {\bbPhi_{\rmD}} }

\def \newenc { {\bbPsi_{\rmE}} }
\def \newdec { {\bbPsi_{\rmD}} }

\def \thetaenc { \bbXi }
\def \thetadec { \bbTheta }

\DeclareMathOperator{\E}{\mathbb{E}}

\usepackage[colorinlistoftodos]{todonotes}
\usepackage{balance}

\colorlet{LightTeal}{white!70!teal}
\colorlet{LightOrange}{white!70!orange}
\colorlet{LightRed}{white!70!red}

\definecolor{DarkGreen}{RGB}{1,100,32}

\usepackage{mysymbol}

\title{Graph Signal Generative Diffusion Models}

\name{Yi\u{g}it~Berkay~Uslu$^{\dagger}$ \quad Samar~Hadou$^{ \dagger}$ \quad Sergio Rozada\sthanks{This work was partially supported by the Spanish AEI (10.13039/501100011033) grant PID2022-136887NB-I00, and the Community of Madrid via the Ellis Madrid Unit and grants URJC/CAM F1180 and TEC-2024/COM-89.} \quad Shirin~Saeedi~Bidokhti$^{\dagger}$ \quad Alejandro~Ribeiro$^{\dagger}$}

\address{$^{\dagger}$Dept. of Electrical and Systems Engineering,
University of Pennsylvania, Philadelphia, PA, USA \\
$^{\star}$Dept. of Signal Processing and Communications, King Juan Carlos University, Madrid, Spain}

\begin{document}
\ninept
\maketitle
\begin{abstract}
We introduce U-shaped encoder-decoder graph neural networks (U-GNNs) for stochastic graph signal generation using denoising diffusion processes. The architecture learns node features at different resolutions with skip connections between the encoder and decoder paths, analogous to the convolutional U-Net for image generation. The U-GNN is prominent for a pooling operation that leverages zero-padding and avoids arbitrary graph coarsening, with graph convolutions layered on top to capture local dependencies.
This technique permits learning feature embeddings for sampled nodes at deeper levels of the architecture that remain convolutional with respect to the original graph.
Applied to stock price prediction---where deterministic forecasts struggle to capture uncertainties and tail events that are paramount---we demonstrate the effectiveness of the diffusion model in probabilistic forecasting of stock prices. 
\end{abstract}


\section{Introduction}
Data defined on irregular domains is pervasive, with applications ranging from recommender systems to wireless networks \cite{ortega2018graph, huang2018rating, he2021overview, chien2024opportunities}.
This paper addresses the problem of generating signals (data) on a given graph when the underlying distribution is unknown and may not admit a tractable form.
To that end, we leverage generative diffusion models \cite{ho2020denoising, song2021denoising}, where the central idea is to gradually corrupt a sample from the original distribution with noise in a forward process until it becomes indistinguishable from a simple, non-informative distribution, e.g. Gaussian.
By reversing this process, the noise can be transformed back into samples of the original distribution.
The backward dynamics rely on a parametric denoiser that removes noise at different noise levels to progressively recover the clean signal \cite{ho2020denoising, song2021denoising}.
While this approach has achieved remarkable success in image \cite{yang2023diffusion, cao2024survey} and graph generation \cite{jo2022score, vignac2023digress}, it has received less attention in the setting where signals are defined on a fixed, given graph.

Graph signal generation has been explored across diverse domains such as recommender systems \cite{chen2024g, zhu2024graph, zhao2024denoising}, spatio-temporal forecasting \cite{wen2023diffstg, zhang2025tsgdiff, hu2024towards}, finance \cite{daiya2024diffstock, you2024dgdnn}, or wireless communications \cite{letafati2024conditional, uslu2025generative, kan2025gdplan}. 
In a nutshell, these works adapt the successful denoising diffusion probabilistic model (DDPM) framework \cite{ho2020denoising} from computer vision  to graphs, parametrizing the backward process with graph neural networks (GNNs) \cite{gama2020graphs, ruiz2021graph}, given that signals are defined on a known graph.
Despite these advances, existing approaches are application-driven, crafting tailored solutions that are usually used to solve specific downstream tasks.
As a result, most methods parametrize the backward process with architectures that blend domain-specific knowledge with standard GNNs \cite{zhu2024graph, wen2023diffstg, hu2024towards} or graph attention mechanisms \cite{zhang2025tsgdiff, daiya2024diffstock}.
However, architecture design should transcend application domains and be guided by the requirements of diffusion. 
Evidence from image generation highlights the importance of this choice, with U-Net \cite{ronneberger2015u} emerging as the canonical design by combining encoder pooling for global context with decoder upsampling and skip connections for fine detail \cite{ho2020denoising}.
A comparable architectural design for graph-based domains is required to establish a general framework for graph-signal diffusion beyond application-specific solutions.

Motivated by this landscape, we propose a general generative modeling framework that extends DDPM for graph signals and unifies the main approaches across application domains.
Furthermore, we introduce a novel architecture that generalizes the U-Net architecture to graph convolutions by introducing a pooling scheme that leverages zero-padding and nested sampling matrices to construct valid graph convolutional features and aggregation neighborhoods supported over the original graph.
To illustrate the framework, we consider stock price prediction, a setting where generative models can capture uncertainties and tail events that deterministic forecasts overlook, and where capturing uncertainties is valuable not only for prediction but also for risk management \cite{feng2016signal}. The summary of the contributions is as follows.
\begin{itemize}[leftmargin=15pt,labelsep=0.5em]
    \item[\textbf{C1}.] We formulate a generative diffusion model for stochastic graph signals that unifies existing application-driven approaches 
    \item[\textbf{C2}.] We propose a novel GNN architecture, termed U-GNN, that leverages graph convolutions and zero-padding graph pooling to capture local dependencies at different resolutions.     
    \item[\textbf{C3}.] We validate our method on stock-price forecasting, where stochastic predictions are vital to represent uncertainties.  
\end{itemize}

\section{Diffusion Models for Graph Signals}
\label{sec:graph-signal-generative-models}

Consider a weighted, undirected graph $\ccalG = (\ccalV, \ccalE, \ccalW)$, where $\ccalV$ is the node set with $|\ccalV| = N$, $\ccalE$ is the edge set, and $\ccalW: \ccalE \mapsto \reals$ is a function assigning weights to edges. We represent graph-structured data with a graph signal $\bbX \in \reals^{N \times F}$, where $F$ is the number of features per node. Equivalently, we use a vectorized form $\bbx \in \reals^{NF}$, obtained by stacking the rows of $\bbX$. Data exchange among nodes is modeled through graph shift operators, $\bbS \in \reals^{N \times N}$, which respect the sparsity pattern of the graph $\ccalG$, i.e., $[\bbS]_{mn} \neq 0$ iff $(m, n) \in \ccalE$ or $m=n$. Common examples of graph shift operators include the adjacency matrix and the graph Laplacian matrix.

We are given i.i.d. samples of a graph signal $\bbx \sim q_0( \cdot \cond \bbS, \bbu)$ supported on a graph $\bbS$ and additional conditional information represented by another graph signal $\bbu$, and we seek to learn a (conditional) generative model $p_{\bbtheta}( \cdot \cond \bbS, \bbu)$ that approximates $q_0( \cdot \cond \bbS, \bbu)$. To this end, we use forward/backward diffusion processes that transport the probability mass between the (conditional) data distribution $q_0$ and a prior distribution $q_T = \ccalN(\mathbf{0}, \bbI)$ over $T$ diffusion time steps. 

The forward diffusion process gradually corrupts data samples $\bbx_0$ by adding Gaussian noise at each step. Given an increasing noise schedule $\{\beta_t\}_{t = 1}^{T}$, e.g., linear, the forward diffusion process follows a Markov process with one-step transition densities,
\begin{align}
q_t(\bbx_t | \bbx_{t-1}) = \ccalN(\bbx_t; \sqrt{1 - \beta_t} \bbx_{t-1}, \beta_t \bbI).
\end{align}
In particular, the process $\bbx_{1:t} | \bbx_0$ is Gaussian and so are the conditional densities $q_t(\bbx_t \cond \bbx_0)$ at all steps $t$. This permits expressing the forward process with a reparametrization \cite{ho2020denoising} as  
\begin{align} \label{eq:forward-process}
    \bbx_t(\bbx_0, \bbepsilon) = \sqrt{ \bar{\alpha}_t } \bbx_0 + \sqrt{1 - \bar{\alpha}_t } \bbepsilon, \quad \bbepsilon \sim \mathcal{N}(\bb0, \bbI).
\end{align}
for $t = 1, \ldots, T$, where $\alpha_t := 1 - \beta_t$ and $\bar{\alpha}_t := \prod_{i=1}^t \alpha_i$. For sufficiently large diffusion time steps $T$, the signal converges to isotropic Gaussian noise, i.e., $\bbx_T \sim \mathcal{N}(\mathbf{0}, \bbI)$. 

The reverse (backward) diffusion process is also a Markov chain that is a time-reversal of the forward process. That is, the process starts from the prior distribution $q_T$, terminates at the (conditional) data distribution $q_0$ and shares the same marginals $\{q_{t} \}_{t = 1}^T$ as the forward process. To this end, the backward diffusion process learns a denoiser, parametrized by $\bbtheta$, that extracts the clean signal $\bbx_0$ from $\bbx_t$, given the conditional information,
\begin{align} \label{eq:ddpm-training}
    \bbtheta^\star \in \argmin_{\bbtheta} \, \E_{\bbx_0, t, \bbu} \left \| \bbx_{\bbtheta} \big( \bbx_t(\bbx_0, \bbepsilon), t; \bbS, \bbu  \big) - \bbx_0 \right \|^2.
\end{align}
\noindent In \eqref{eq:ddpm-training}, the expectation is over the joint distribution of conditional information, graph data samples, and diffusion time steps sampled uniformly in $[1, T]$. Alternatively, DDPM training \cite{ho2020denoising} learns the parametric function $\bbepsilon_{\bbtheta^*}(\bbx_t, t; \bbS, \bbu)$ that approximates the noise $\bbepsilon$ added to $\bbx_0$ at each time step $t$,
\begin{align} \label{eq:ddpm-noise-training}
    \bbtheta^\star \in \argmin_{\bbtheta} \, \E_{\bbx_0, t, \bbu} \left \| \bbepsilon_{\bbtheta} \big( \bbx_t(\bbx_0, \bbepsilon), t; \bbS, \bbu  \big) \!-\! \bbepsilon \right \|^2\!.
\end{align}

Given an optimal noise predictor $\bbepsilon_{\bbtheta^\star}$, the backward diffusion process reverts the forward noising process in \eqref{eq:forward-process} at each  step by,
\begin{align} \label{eq:backward-diffusion-sample}
    \bbx_{t-1} = \frac{1}{\sqrt{\alpha}_t} \Big( \bbx_t - \frac{\beta_t}{\sqrt{1 - \bar{\alpha}_t}} \bbepsilon_{\bbtheta^\star} \big( \bbx_t, t; \bbS, \bbu \big) \Big) + \sqrt{\beta_t} \bbw,
\end{align}
where $\bbw \sim \mathcal{N}(\bb0, \bbI)$. Put together, generating novel graph signal samples approximately distributed by the conditional data distribution, i.e., $\bbx_0 \sim p_{\bbtheta}(\cdot \cond \bbS, \bbu) \approx q_0(\cdot \cond \bbS, \bbu)$, boils down to learning an optimal graph signal denoiser [cf.~\eqref{eq:ddpm-training}], or equivalently, an optimal noise predictor [cf.~\eqref{eq:ddpm-noise-training}], and running the sampling equation \eqref{eq:backward-diffusion-sample} iteratively for $t = T, \ldots, 1$ with $\bbx_T \sim p_T$. Consequently, we leverage a graph neural network (GNN) architecture for the noise parametrization $\bbepsilon_{\bbtheta}$, which we present next.

\section{U-Graph Neural Networks}
\label{sec:gnn_architecture}

We introduce a graph neural network (GNN) architecture for the parametrization of $\bbepsilon_{\bbtheta}$, dubbed U-Graph Neural Network or U-GNN for brevity. A U-GNN with depth $B$ consists of $2B$ GNN blocks arranged in a U-shape with a left and right path, mirroring convolutional U-Net architectures. 

The left path is an encoding (contraction) path and comprises $B$ encoding GNN blocks, which successively extract graph convolutional features and downsample the signals. The right path is a decoding (expansion) path and also consists of $B$ decoding GNN blocks, which upsample the graph convolutional features and combine them with higher-resolution information provided by skip connections from the left encoding path. Next, we describe a GNN architecture that forms the core of the encoding and decoding blocks. 

\subsection{Graph Neural Networks}
GNNs learn local graph convolutional features by successive applications of graph filters and pointwise nonlinearities~\cite{gama2018convolutional}. A GNN is a parametrized map $\bbPhi$ that, given a shift operator $\bbS$, transforms an input signal $\bbX \in \reals^{N \times F_\text{in}}$ into an output signal $\bbV \in \reals^{N \times F_\text{out}}$,
\begin{align} \label{eq:gnn}
    \bbV = \bbPhi\big(\bbX, \bbS; \ccalH \big),
\end{align}
where $\ccalH$ denotes a set of learnable parameters. The map $\bbPhi$ is a linear composition of $L$ graph convolutional layers, defined as
\begin{align} \label{eq:graph-convolution-layer}
    \bbV_\ell =  \varphi \Bigg(\, \sum_{k = 0}^{K_{\ell}}  \bbS^k \, \bbV_{\ell-1}  \bbH_{k,\ell}  \,\Bigg), 
\end{align}
with $\bbV_L=\bbV$. The core component of each layer is a polynomial graph filter that aggregates features over $k$-hop neighborhoods up to $K_\ell$ via the learnable parameters $\{\bbH_{k,\ell} \}_{k=0}^{K_\ell}$. The filter is then followed by a pointwise nonlinear activation function $\varphi$, e.g., ReLU. Additional operations such as a normalization layer, e.g., batch or layer normalization, and a pooling operation, e.g., max-pooling, can be subsumed in the definition of $\varphi$.

Preceding the first convolutional layer, we incorporate time and conditional embeddings into the input signal $\bbX$ via addition and concatenation, respectively. That is, the input to the first graph filter is
\begin{align}
    \bbV_0 = \big[ \bbUpsilon_\rmX(\bbX) + \bbUpsilon_\rmT(\bbt) ; \, \bbUpsilon_{\rmU}(\bbU) \big],
\end{align}
where $\bbUpsilon_\rmT: \reals^N \mapsto \reals^{N \times (F_\text{in}/2)}$ is a sinusoidal embedding of vectorized diffusion time steps $\bbt = t \mathbf{1}_{N}$. The learnable maps $\bbUpsilon_{\rmX}: \reals^{N \times  F_\text{in}} \mapsto \reals^{N \times (F_\text{in}/2)}$ and $\bbUpsilon_{\rmU}: \reals^{N \times  U} \mapsto \reals^{N \times (F_\text{in}/2)}$ project the input signal $\bbX$ and the conditional signal $\bbU$ into $F_\text{in}$-dimensional  embeddings, respectively, with parameters included in $\ccalH$. Note that $t$ and $\bbU$ are inputs to the GNN, though omitted from the representation in \eqref{eq:gnn} for brevity.

\subsection{Encoding \& Decoding Blocks}

The operation of both encoding and decoding blocks is captured by the GNN defined in \eqref{eq:gnn}. Along the encoding path, the $F_{b-1}$-dimensional feature signal $\bbX_{b-1}$, produced by the $(b - 1)$th block, is passed to the $b$th block to generate
\begin{align} \label{eq:encoder-block}
    \bbX_b = \enc \big( \bbX_{b - 1}, \bbS; \thetaenc_b  \big),
\end{align}
where $\thetaenc_b$ is the set of learnable parameters of the $L$-layered GNN in the $b$th encoding block. The encoding path composes \eqref{eq:encoder-block} for $b = 1, \ldots, B$ to obtain the sequence of graph convolutional features $\{ \bbX_b \}_{b = 1}^{B}$ with reduced dimensionality $F_{B} < F_{B-1} < \dots < F_0$. The input feature signal $\bbX_0 = \bbPi(\bbX)$ is generated by a read-in map $\bbPi: \reals^{N \times F} \mapsto \reals^{N \times F_0}$, where $\bbX$ is the matrix representation of the given input $\bbx_t \in \reals^{NF}$ to $\bbepsilon_{\bbtheta}$.

The decoding path traverses the opposite direction to the encoding path and upsamples the low-dimensional features. The inputs to a decoding block at depth $b$ are the output of the decoding block at depth $b+1$ and the skip connection from the encoding block at depth $b$.
The decoding block concatenates the inputs in the feature dimension and processes them with a GNN, 
similar to the encoding blocks, as
\begin{align} \label{eq:decoder-block}
\bbY_{b - 1} = \dec \Big( \big[ \bbY_{b}; \bbX_b \big], \, \bbS; \, \thetadec_b \Big),
\end{align}
where $\thetadec_b$ is the set of learnable parameters of the $L$-layered GNN in the $b$th decoding block. For the decoding block at depth $B$, $\bbY_B$ is obtained by a dimension-preserving multilayer perceptron (MLP) transformation of the encoder final output $\bbX_B$. The output of the decoding block at depth one $\bbY_0$ is finally mapped by a read-out layer to the model target.

\subsection{Down- \& Upsampling with Sampling Matrices}
The encoder-decoder architecture described so far performs dimensionality reduction only in the feature dimension. However, this need not be the case. Each encoding block can be augmented to pair the GNN
with a graph pooling (node downsampling) operator, resulting in output features $\bbX_b \in \reals^{N_b \times F_b}$ for all $b$, with $N_b\leq N_{b-1}$ and $N_0=N$ for notational consistency. Similarly, decoding blocks can symmetrically upsample in both node and feature dimensions. Except the following challenge arises after the first encoder block. 

The input signal to the first encoder block $\bbX_0 \in \reals^{N_0 \times F_0}$ is already supported on the original graph $\bbS$ with $N_0$ nodes, and therefore graph convolutions at this depth are straightforward. However, the subsequent pooling operations leave us with a downsampled node signal $\bbX_{b} \in \reals^{N_b \times F_b}$, which is lower dimensional than the input signal and is no longer supported on the original graph $\bbS$. 

To resolve this support mismatch, we treat downsampling as a node selection operation from a predefined subset and leverage \emph{zero padding} to lift downsampled signals back to the original node set. Formally, we represent the node selection mechanism at depth $b$ with a binary fat selection matrix $\bbC_{b} \in \{0, 1\}^{N_{b} \times N_{b - 1}}$ that satisfies $\bbC_{b} \mathbf{1}_{N_{b - 1}} = \mathbf{1}_{N_{b - 1}}$ and $\bbC^\top_{b} \mathbf{1}_{N_{b}} \preceq \mathbf{1}_{N_{b}}$.  Then, the product $\bbC_{b} \bbX_{b-1}$ selects $N_{b}$ distinct nodes out of the $N_{b - 1}$ rows of $\bbX_{b-1}$, whereas $\bbC^\top_{b} \bbX_{b}$ lifts $\bbX_b$ back to the $N_{b-1}$-node domain by inserting zeroes at the indices of the nodes dropped during downsampling.

More generally, we track the nodes selected from the original graph at depth $b$ by a nested sampling matrix $\bbD_{b}\in \{0,1\}^{N_b \times N_0}$,
\begin{align} \label{eq:nested-sampling-matrix}
\bbD_{b} = \bbC_{b} \bbC_{b - 1} \cdots \bbC_1 = \prod_{i = 1}^{b} \bbC_i.
\end{align}
Here, the product $\widetilde{\bbX}_{b} = \bbD^\top_{b} \bbX_{b}$ zero pads $\bbX_{b}$ such that the features of the active nodes remain intact at their original indices with respect to $\bbS$, while the features of the inactive nodes are set to zero. Hence, the signal $\widetilde{\bbX}_{b}$ is supported on the original graph $\bbS$, and its convolutions with respect to $\bbS$ are well-defined.

\subsection{Downsampling Encoder \& Upsampling Decoder Blocks}
To incorporate the downsampling and upsampling into the encoder and decoder blocks, respectively, we redefine the GNN map in \eqref{eq:gnn} to accept the nested selection matrix $\bbD$ as an additional input, 
\begin{align}
    \bbV = \bbPsi \big( \bbX, \bbS, \bbD; \ccalH \big).
\end{align}
Accordingly, we rewrite the encoder \eqref{eq:encoder-block} and decoder \eqref{eq:decoder-block} blocks at a given depth $b$ as
\begin{align} 
    \bbX_b &= \newenc \big( \bbC_b \bbX_{b-1}, \, \bbS, \, \bbD_b; \, \thetaenc_b \big), \label{eq:new-encoder-block} \\
    \bbY_{b - 1} &= \newdec \left( \bbC^\top_b \big[\bbY_b ;\, \bbX_b \big], \, \bbS, \, \bbD_{b-1}; \,  \thetadec_{b} \right). \label{eq:new-decoder-block}
\end{align}
Each block uses the selection matrix corresponding to the resolution of its output, which is $\bbD_b$ for the encoder and $\bbD_{b-1}$ for the decoder.
This incorporation of the nested selection matrix transforms the cornerstone operation of the GNN, i.e., the graph convolution \eqref{eq:graph-convolution-layer}, to
\begin{align} \label{eq:proposed-graph-conv-layer}
 \bbV_{\ell} = \varphi \left( \,  \sum_{k = 0}^{K_{\ell}}  \left[ \bbD \big( \bbS^{\gamma} \big)^k \bbD^{\top} \right] \bbV_{\ell - 1} \bbH_{k, \ell} \, \right),
\end{align}
where $\gamma$ is a positive integer that determines the stride of the graph shift operation, and its purpose will be clarified shortly. The rest of the GNN architecture remains the same. We note that for $\gamma = 1$ and $\bbD = \bbI$, \eqref{eq:proposed-graph-conv-layer} coincides with \eqref{eq:graph-convolution-layer}.

The convolution in \eqref{eq:proposed-graph-conv-layer} is noteworthy for offering two computationally distinct yet mathematically equivalent viewpoints. The first exploits sampling directly by applying the $k$ reduced shift operators $\bbD \big( \bbS^\gamma \big)^k \bbD^\top$ to the subsampled input signal. This approach avoids redundant computations on inactive nodes but requires computing the low-dimensional shift matrices for each $k$ beforehand. The second entails first zero-padding the downsampled signal up to an $N$-dimensional signal, followed by repeated application of the powers of $\bbS$ to obtain $\gamma k$-shifted versions, and lastly subsampling to the desired resolution. We realize this perspective with sparse matrix multiplications for a low computational cost in our implementation.

Finally, the stride parameter $\gamma$ controls the effective hop size of the graph shift operator. In the zero-padded domain, where $\gamma > 1$, $\bbS^\gamma$ redefines the $k$-hop aggregation neighborhoods, so that we can choose larger neighborhoods that are effectively intersected with the active nodes selected by $\bbD$. This technique remedies the issue of aggregating and pooling mostly zeroes at deeper levels of the U-GNN when zero-padded signals are highly sparse due to successive downsampling operations at higher levels.

\section{Case Study:
Stock Price Forecasting}

Consider a stock market with $N$ stocks. In stock price forecasting, we are interested in predicting the evolution of log returns over a future horizon of $T_\rmh$ days $\{ \bbx^{(T_\rmp + 1)}, \ldots, \bbx^{(T_\rmp + T_\rmh)} \}$, given a history of past observations over the past $T_\rmp$ days $\{ \bbu^{(1)}, \ldots, \bbu^{(T_\rmp)} \}$. Here, $\bbx^{(k)} \in \reals^N_+$ denotes the daily log return ratio of stock prices, whereas $\bbu^{(k)} \in \reals^{N \times U}$ is a collection of $U$ quantities of interest (stock features), including the stock prices themselves, observed on day $k$.

We are also provided with certain long-term financial indicators (stock fundamentals) such as sector information, market capitalization, trailing price-to-earnings (P/E), profit margins, and so on. From these indicators, we compute a normalized correlation matrix $\bbSigma$.
We model the stock relationship with a graph, and  let the weighted adjacency matrix $\bbS$ be the correlation matrix with zeros on the diagonal.
 
Our goal is to learn a generative model that samples future realizations (trajectories) of log returns $\bbx = [ \bbx^{(T_\rmp + 1)} , \ldots, \bbx^{(T_\rmp + T_\rmh)}]$ given a graph representation $\bbS$ of stock relationships and a conditional graph signal $\bbu = [ \bbu^{(1)}, \ldots, \bbu^{(T_\rmp)} ]$ pertaining to certain stock features observed over the most recent $T_\rmp$ days. Here we stacked the temporal signals along the feature dimension and wrote them in vector form to match notation with the U-GNN architecture interface.

\begin{figure*}[t]
    \centering
    \begin{minipage}{.24\linewidth}
        \includegraphics[height=3.4cm,width=\linewidth]{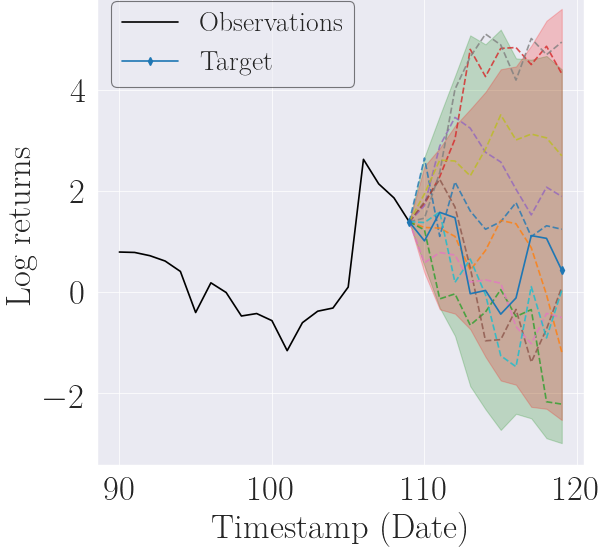}
    \end{minipage}
    \begin{minipage}{.24\linewidth}
        \includegraphics[height=3.4cm,width=\linewidth]{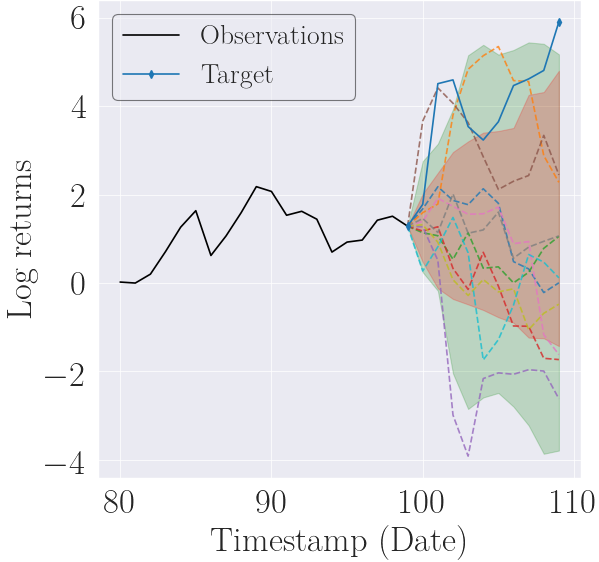}
    \end{minipage}
    \begin{minipage}{.24\linewidth}
        \includegraphics[height=3.4cm,width=\linewidth]{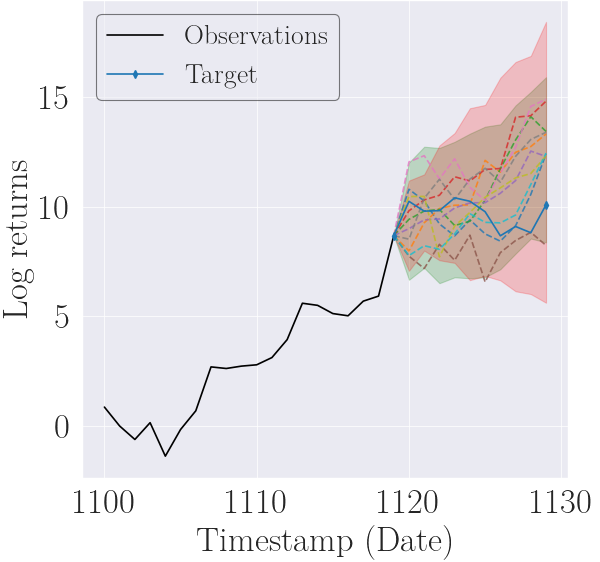}
    \end{minipage}
    \begin{minipage}{.24\linewidth}
        \includegraphics[height=3.4cm,width=\linewidth]{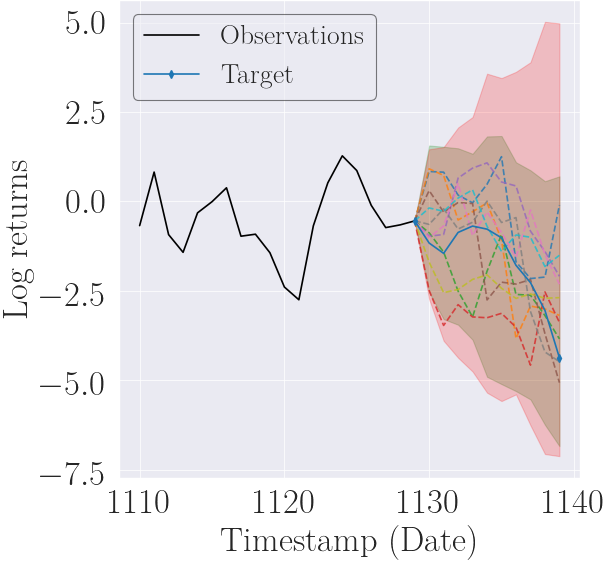}
    \end{minipage}
    \caption{Ten stochastic log-return forecasts over a $T_\rmh = 10$-day horizon (dashed), conditioned on a $T_\rmp = 20$-day history (black) for $4$ stocks. For clarity, we plot only 10 trajectories generated by U-GNN. Green and light red shadowed bands cover the 95\% confidence intervals for U-GNN and GRW trajectories, respectively. The realized path generally lies within the ensemble of sampled U-GNN trajectories.}
    \label{fig:prediction}
\end{figure*}

\begin{table*}[ht!]
    \centering
    \caption{Comparison of U-GNN against GRW baseline across several configurations and error metrics. A lower value is better for all metrics.}
    \label{table:error}
    \vspace{-.7em}
    \addtolength{\tabcolsep}{-0.21em}
    \begin{tabular}{ccccccccccccccccc}
    \toprule
    \multirow{2}{*}{}   & \multicolumn{4}{c}{ $T_\rmp = 20$, $T_\rmh = 10$} 
                        & \multicolumn{4}{c}{ $T_\rmp = 20$, $T_\rmh = 20$  } 
                        & \multicolumn{4}{c}{ $T_\rmp = 5$, $T_\rmh = 5$  } 
                        & \multicolumn{4}{c}{ $T_\rmp = 10$, $T_\rmh = 1$  } \\
    \cmidrule(lr){2-5} \cmidrule(lr){6-9} \cmidrule(lr){10-13}  \cmidrule{14-17}   
                        & MIS & CRPS & RMSE & MAE
                        & MIS & CRPS & RMSE & MAE
                        & MIS & CRPS & RMSE & MAE
                        & MIS & CRPS & RMSE & MAE \\
    \midrule
    U-GNN    & 20.6 & \textbf{1.09} & \textbf{2.25} 
                & 1.74 & 30.8 & \textbf{1.61} & \textbf{3.14} & \textbf{2.28} & 35.0 & \textbf{1.33} & \textbf{2.20} & \textbf{1.50} & 23.9 & 0.85 & 1.28 & 0.93 \\
    GRW           & \textbf{17.1} & 1.32 & 2.39 & 1.76 &                 30.6 & 1.98 & 3.56 & 2.61 &                 \textbf{29.7} & 1.62 & 2.98 & 2.08 & 
                    \textbf{11.2} & 0.73 & 1.29 & 0.95 \\
    \bottomrule
    \end{tabular}
\end{table*}

\subsection{Training Details}
We experiment on the S\&P100 dataset, which includes stocks of $N = 100$ U.S. companies from diverse industries with top market capitalization. We obtain the historical values of the stocks over the last $5$ years using the open-source ``yfinance'' library. Given $T_\rmp$ and $T_\rmh$ values, we split the 5 years period into moderately-sized chunks, shuffled them, and performed a $90:5:5$ split across training, validation, and test datasets. We use the validation dataset solely for early stopping and report the performance on the test dataset. 

In our main experiment configuration, the past and future window lengths are $T_\rmp = 20$ and $T_\rmh = 10$, respectively. The U-GNN has depth $B = 3$ and each GNN block itself contains $L = 2$ layers with $K_{1} = K_2 = 2$ filter taps and the stride is $\gamma = 1$. We set the initial number of hidden feature channels to $F_0 = 64$, which is halved at each depth of the encoding path. We skip the downsampling operation in the first level of U-GNN and downsample twice by a factor of $N_3 / N_2 = N_{2} / N_{1} = 0.8$ in the remaining two levels. We adopted a node-degree based selection approach for its simplicity, where we always dropped nodes with the least degree in the graph.

The diffusion process is run for a total of $T=500$ time steps and a cosine noise schedule $\{ \beta_t \}_{t = 1}^{T}$ is used. We train the U-GNN over mini-batches of size $64$ for a maximum of $5 \times 10^4$ epochs with an AdamW optimizer and an initial learning rate of $2 \times 10^{-2}$ that decays according to a cosine schedule with warm restarts. 

\subsection{Evaluation Metrics \& Test Performance}
In line with the assumption of log-normality of stock prices adopted in most financial models, we benchmark U-GNN against a baseline that models the future evolution of log returns of a stock by a geometric random walk (GRW) whose drift and diffusion coefficients are estimated from the past window. 

For evaluation, we sample 20 trajectories of length $T_\rmh$ for each stock from both U-GNN and GRW. 
We report the root-mean-squared error (RMSE) and mean absolute error (MAE) metrics between the ground truth (target) trajectory and a mean trajectory obtained by ensemble averaging. We also quantify the fidelity of the distribution of trajectories with two probabilistic metrics.
Continuous ranked probability score (CRPS) \cite{matheson1976scoring} measures the discrepancy between the predicted cumulative distribution function (CDF) of a forecast and a given observation, whereas mean interval score (MIS) evaluates the confidence of a model in its prediction intervals \cite{gneiting01032007}.

Fig.~\ref{fig:prediction} shows (cumulative) log-return forecasts for four different stocks over a $10$-day horizon conditioned on a $20$-day history. We plot ten trajectories sampled using our trained U-GNN via \eqref{eq:backward-diffusion-sample} to capture predictive uncertainty and tail risk. As shown in the figure, the realized path (blue) typically lies within the envelope of the drawn samples and aligns with their dominant direction, indicating the model's ability to produce coherent stochastic scenarios. For example, in the leftmost panel, the conditional window trends very slightly downward for roughly $10$ days before tilting up sharply and then down again, and most sample trajectories pivot downward, tracking the ground truth. However, a minority of samples continue an upward move, reflecting residual uncertainty and a nonzero probability of a continued drawup. The remaining panels display less volatile scenarios. Overall, U-GNN balances accuracy with uncertainty robustness better than a GRW baseline.

In Table~\ref{table:error}, we report comparative results for four different experiment setups, including the main configuration. We observe that the U-GNN either significantly outperforms or fares similarly to the baseline across all setups and all three evaluation metrics, except for the MIS. We note that this is consistent with the MIS scoring the confidence in prediction intervals, and with the random walk placing the bulk of its probability mass around its mean. The generative model of stock prices might not be as confident in typical scenarios but is capable of capturing rare events and volatilities that are critical to decision-making, as is the case in financial markets. Node pooling (downsampling) did not produce sizable gains in our experiments. We expect more substantial gains on larger graph datasets. Furthermore, better-performing U-GNN models can be trained, specifically for time-series forecasting, by taking into account the spatial and temporal dependencies together, which we leave as future work.

\section{Conclusion}
In this work, we proposed a graph signal generative modeling framework for sampling stochastic graph signals over given graphs and conditional graph signals. We introduced a novel diffusion model architecture termed U-GNN that leverages GNNs as core operational blocks and a graph pooling technique in an effort to generalize convolutional U-Nets. We applied the proposed methodology and U-GNN architecture to a stock price prediction problem. Future work will include a more comprehensive study of the U-GNN architecture.


\clearpage
\newpage
\balance
\bibliographystyle{IEEEbib}
\bibliography{strings,refs}

\end{document}